\documentclass{article}
\usepackage[utf8]{inputenc}
\usepackage{graphicx}
\usepackage{amsthm}
\theoremstyle{definition}

\usepackage{amsmath}
\usepackage{subcaption}

\usepackage[dvipsnames]{xcolor}

\newcommand{\C}{{\mathcal C}}

\title{A New Arc-Routing Algorithm Applied to Winter Road Maintenance \footnote{This research is conducted within the project Network Optimization (17-10090Y) supported by Czech Science Foundation.}}
\begin{document}
\maketitle
\begin{center}
{\bf Jiří Fink}\\ Department of Theoretical Computer Science and Mathematical Logic, Charles University, Czech Republic

\smallskip
{\bf Martin Loebl}\\ Department of Applied Mathematics, Charles University, Czech Republic

\smallskip
{\bf Petra Pelikánová}\\ Department of Applied Mathematics, Charles University, Czech Republic   
\end{center}

\section*{Abstract}
This paper studies large scale instances of a fairly general arc-routing problem as well as incorporate practical constraints in particular coming from the scheduling problem of the winter road maintenance (e.g. different priorities for and methods of road maintenance). We develop a new algorithm based on a bin-packing heuristic which is well-scalable and able to solve road networks on thousands of crossroads and road segments in few minutes. Since it is impossible to find an optimal solution for such a large instances to compare it with a result of our algorithm, we also develop techniques to compute lower bounds which are based on Integer Linear Programming and Lazy Constraints.

\section{Introduction}
Our goal is to develop an algorithm for the Arc Routing Problem (ARP) which is able to handle large scale instances and incorporate practical constraints, i.e., different priorities and methods of road maintenance.

\subsection{Description of the problem}

We study a variant of arc routing problem
\begin{itemize}
\item with \textbf{multiple cars} and every road of given length is maintained by exactly one car (deadheads are allowed but have to be minimized);
\item with \textbf{multiple depots} and every car has to start and finish in the same depot;
\item with a given \textbf{deadline} in which all roads have to be maintained; and
\item every road has to be traversed in \textbf{both direction} by the same car (one-way roads are not considered).
\end{itemize}

\noindent In our model, every road has assigned one of three following {\bf priorities}:
\begin{enumerate}
    \item Highways and interstate roads
    \item Intercity roads and roads used e.g. by bus routes
    \item Low-important roads connecting villages to towns or intercity roads
\end{enumerate}
The exact role of priorities is discussed in Section \ref{sec:model}. Actually, our approach can be generalized for arbitrary many priorities.

Furthermore, every road has assigned one of the following method of maintenance
\begin{description}
    \item[Chemical] (e.g. sodium chloride, calcium chloride, brine): This is a preferred method especially for 
    roads with heavy traffic.
    \item[Inert] (i.e. sand, brash, slag):  Since it is prohibited to spread chemical materials in nature parks and in protected areas around water sources, inert materials is used there.
    \item[Snowplow] Some roads with very low traffic, especially in mountain areas where vehicles have to use snow chains, are served by snowplowing only.
\end{description}
Roads without any form of winter maintenance are excluded from our model. We consider two types of cars, one for chemical and the other for inert maintenance; and both are equipped with a blade for snowplowing. Mathematical model of our problem is stated in Section \ref{sec:model}.

The primal objective in our problem is minimizing the number of cars, and the secondary objective is minimizing deadhead; i.e. traversing without maintenance.

\subsection{State of the art}
An overview of literature on the problem of winter road maintenance and its solutions is \cite{PLC1, PLC2, PLC3, PLC4}. An excellent recent overview illustrating main works on the General Routing Problem can be found in \cite{BLMV} where the authors design a new branch-and-cut algorithm for the capacitated general routing problem. In \cite{PLA}, the authors address the snow plowing problem where the area to be maintained is partitioned into fixed sub-areas. The maintaining vehicles are not allowed to cross the boundaries of the sub-areas. This is a common administrative restriction which may worsen the solution; in our considerations we avoid it. The authors also consider road priorities and a precedence relation between roads of different priority. We maintain these conditions in our model.
In \cite{KHS}, the authors aim at constructing the routes schedule minimising the longest route; the network may have one-way streets and is modelled as a mixed graph.

\medskip\noindent

In most of the literature (see also  \cite{BYL}, \cite{GMH} or \cite{GB} for recent works not covered by \cite{BLMV}) relatively small scale problems are considered with, e.g., at most 100 road segments, with only one source-sink networks. The usual approach is an integer programming formulation solved by a linear programming relaxation accompanied with a heuristics. Also, only one source-sink is considered usually. 

Kinable et.al. \cite{KHS} study a real-world snow plow routing problem (in the USA) and they compare three methods based on Integer Linear Programming (LP), Constrain Programming (CP) and a local heuristic. The LP model uses more than $C E^2$ integer variables where $C$ and $E$ is the number of cars and roads segments, respectively. If this method is applied to the instance reported here in the case-study section, the number of variables would be more than 500 millions which current computers cannot handle; this confirms authors conclusion that LP can be used only for small instances. CP can handle large instances and find good solutions to instances up to a 1000 road segments, but does not scale well beyond that. The third method is based on a greedy construction of a feasible initial solution followed by an acceptance improvement heuristic. The heuristic utilizes two simple search neighborhoods: bestSwapMove which swaps two jobs (cleaning a road or refuelling) of two cars, and bestRemoveInsertMove which moves a job from a car to another one. In this paper, we develop significantly stronger heuristic which essentially swaps arbitrary subset of jobs between two cars.

Ciancio et.al. \cite{CDV} applied Branch-price-and-cut method for the Mixed Capacitated General Routing Problem with Time Windows. They tested their method for graphs up to 380 edges which is smaller than our network. 

\subsection{Main contribution}

Main contributions of this paper are the following.

\begin{itemize}
\item Introduce a new variant of arc routing problem which includes additional real-life constrains.
\item Develop a heuristic algorithm which works well on large-scale case studies and it is scalable even for huge road networks.
\item Introduce lower bounds for optimal solutions.
\item The implementation solves a realistic road network with thousands of roads in few minutes on an ordinary laptop, so that it can be used as a subroutine e.g. for real-time control.
\end{itemize}
    
\section{Model of the winter road maintenance} \label{sec:model}

A {\it road network} is given as a simple connected graph $G$ with roads (edges) $E$ and crossroads (vertices) $V$. Our method can also handle loops and parallel edges but it prevents directed edges since every road has to be traversed in both directions. For a set of vertices $R \subset V$ we denote by $G[R]$ the subgraph of $G$ induced by $R$, and $\delta_G R$ the set of edges between $R$ and the complement $V \setminus R$.

Since we consider a variant of ARP with multiple cars, we denote a set of cars by $\C$. A set of roads traversed by a car $c \in \C$ is called a {\it plan} and it is denoted by $T_c \subseteq E$. It is easy to observe that there exists a tour passing every road of $T_c$ exactly once in both directions if and only if $T_c$ is connected. Note that such a tour is always closed. Therefore, we require $T_c$ to be connected for every $c \in \C$. 
%On the other hand, we do not need to study such tours since they can be trivially found for a given connected graph in linear time. 
Note that for the connectivity condition it may be necessary to include deadheads in a solution which means that some roads are traversed by more than one car. 

We also have to ensure that every road is maintained by a car and to incorporate different maintenance methods. We require that every chemically maintained road is traversed by a chemical car and every road maintained by inert is traversed by a car spreading inert. Since both types of cars are equipped with a blade, snowplowed roads can be traversed by a car of either type. Let $E_c$, $E_i$, and $E_s$ be the set of road maintained by chemicals, inert and snowplow, respectively. Furthermore for loading maintaining material, a set of depots is given and every depot contains a storage with chemical or inert material (or both). Every car has to pass a depot with the appropriate storage and we denote by $d_c \in V$ the depot of a car $c$.

Every road $e \in E$ has a length $l_e$ and the sum of lengths of roads traversed by each car has to be at most a given limit $L$. This limit and lengths may have various meanings in practice. One may consider that $l_e$ is the time to traverse road $e$ in both directions and $L$ is a deadline in which all roads has to be maintained. Other interpretation is that $l_e$ is the amount of material spread on the road $e$ and $L$ is the capacity of a car. Our method can be easily modified to consider different limits for maintaining methods. By careful choice of $L$ we can assume that loading time and driver's breaks are excluded from the deadline $L$ and we do not incorporate this time in our mathematical model.

Next, we incorporate priorities of roads into our model which are introduced to ensure that more important roads receive better maintenance. However, this goal may be realized in various ways and different legislation is used across the world. One may require that more important roads are maintained several times during the deadline, but this condition can be easily incorporated by increasing the length $l_e$ by the appropriate multiplicity. Others may require that more important roads are maintained first. Furthermore for extreme weather conditions, it would be useful to remove low priority roads from our maintaining plans $T_c$ and concentrate all efforts only on critical roads. Therefore in order to meet these demands, we require that for every car $c \in \C$ the following {\it monotonic property} is satisfied: 
For every road $e \in T_e$ there exists a path in $T_e$ from the depot $d_c$ to $e$
such that priorities (excluding deadheads) along this path are non-decreasing.

This monotonic property ensures that after removing low priority roads from every plan $T_c$, all plans remain connected and contain their original depots, and therefore, there exists a tour starting in a depot and traversing crucial roads without wasting time on remaining roads. In practice, it may not be possible to place a depot directly on a highway, so maintaining cars may have to use local roads to reach a highway. Hence, we do not consider deadheads in the monotonic property.

A set of cars $\C$ and a set of their plans $T = \{T_c;\; c \in \C\}$ form a \emph{solution} of our problem. This solution is \emph{feasible} if all conditions discussed in this section are satisfied. Recall that primal objective in our problem is minimizing the number of cars, and the secondary objective is minimizing deadhead.

\section{Theoretical background}

Our problem is NP-hard, and there are two essential theoretical reasons why we cannot hope to find an optimal solution for large-scale instances in reasonable time: Steiner trees and Bin packing.

\subsection{Steiner trees} \label{sec:steiner}
In the classical version of Steiner tree problem, a graph $G$ with weighted edges and a subset of vertices $S$ are given and the problem is to find a connected subgraph of $G$ containing all vertices of $S$ with the minimal weight. This problem is well-known to be NP-hard \cite{GJ}. For our purposes, an edge version is important: Given a subset of edges (instead of vertices), find a connected subgraph of $G$ containing all prescribed edges with the minimal weight.

This edge version of Steiner tree problem makes our problem hard even if the limit $L$ is sufficiently large to ensure that only one inert (or chemical) car can maintain all inert roads. In this case, the set of all inert roads may be disconnected so a set of chemical roads has to be added to the inert car's plan to ensure connectivity. Finding such a set of chemical roads with the minimal weight is exactly the edge version of the Steiner tree problem. This kind of reasoning is used in Section \ref{sec:lower} to obtain lower bounds on a minimal deadhead.

\subsection{Bin packing}
Bin packing problem is the optimization problem of using minimal number of bins of a given capacity for packing objects of a given variable volume. By analogy, bins are the vehicles and we want to minimize the number of them for servicing the edges of the graph of the road network.
        
Bin packing is a basic NP-hard problem and thus it is usually solved by one of many heuristics designed to solve it \cite{GJ}. Since instances of bin packing problem which we need to solve in our algorithm contain (relatively) small number of roads, we use a brute-force exponential-time algorithm which enumerates all feasible combinations.

\section{Our algorithm}

Our algorithm starts by finding an initial feasible solution which is gradually improved by local changes. Note that after every local change, the solution is always feasible.

\subsection{Initial feasible solution}

It is easy to find an initial feasible solution as follows: For every road $e \in E$ create a new car $c$ of the same maintaining type as $e$ and find the shortest path $T_c$ from $e$ to the nearest depot with the appropriate storage. Theoretically, this solution may not be feasible if there exists a road which is too far from any depot, but then there is no feasible solution. Therefore, we assume that our initial solution is feasible. However, this initial solution is very inefficient, so in our implementation of the algorithm, we added a heuristic which greedily expand every plan as much as the limit $L$ allows before creating a new car.

\subsection{Local changes based on the Steiner tree problem}

The goal is to try to reduce deadheads of a single car $c$. We consider all roads of $T_c$ not traversed by any other car and extend them into a connected graph containing a depot using some heuristic for the Steiner tree problem. Our implementation finds shortest paths between components while preserving the monotonic property. The advantage of this heuristic is simplicity and computation speed. 

\subsection{Local changes based on Bin packing}

Consider two cars $c_1$ and $c_2$ with the same maintaining method traversing roads $T_1$ and $T_2$. If $T_1 \cup T_2$ can be traversed by only one car, then we reduce the number of cars. So, assume that only one car is insufficient. Our goal is to find new plans $T'_1$ and $T'_2$ so that the new solution remains feasible and the weight of $T'_2$ is as small as possible.

We can apply local heuristics by Kinable et.al. \cite{KHS} which try to move one road from $T_2$ to $T_1$ or swap one road of $T_1$ with one road of $T_2$. We can also exchange two roads from $T_1$ with one road from $T_2$, or exchange three roads by two roads, etc. We generalize this idea by exchanging arbitrary subsets of roads.

The basic idea is to use a heuristic for the Bin packing problem to split $T_1 \cup T_2$ into $T'_1$ and $T'_2$ such that the weight of $T'_1$ is as large as possible while it is below the limit $L$. In addition, we have to ensure feasibility, especially connectivity and the monotonic property of both plans $T'_1$ and $T'_2$. Therefore, we enumerate only a subset of all combinations of splitting which ensures feasibility.

For simplicity, we assume that both plans $T_1$ and $T_2$ are trees rooted in depots. In our implementation, we use depth-first search algorithm to find a tree representation of plans and verify that the monotonic condition is satisfied. We call \emph{branches} of a tree $T$ rooted in vertex $u$ all components of the graph $T \setminus u$ that do not contain the depot of $T$.

Let us consider a vertex $u$ shared by plans $T_1$ and $T_2$. The tree $T_2$ may have branches rooted in $u$ which can be maintained by the car $c_1$ while preserving the monotonic condition. Similarly, some branches of $T_1$ rooted in $u$ may be maintained by $c_2$. So, we select all branches rooted in $u$ which can be exchanged between $c_1$ and $c_2$. Using methods for Bin packing problem, we choose such exchange of these branches that minimizes the length of the plan of car $c_2$. In our implementation, we analyze all possible exchanges of branches rooted in all common vertices of $T_1$ and $T_2$ in a single Bin packing instance.

\subsection{Global algorithm}

Using Bin packing, we create some cars with large plans and other cars with small plans. We would like cars with small plans to be close to each other so that we can join them together. To achieve that, we apply the Bin packing approach so that cars with large plans are moved on one 'side' of the network and cars with small plans on the other 'side'. The exact meaning of a "side" is various.

We can use the geographical coordinates of all roads and move large plans e.g. to the north. In this case, we run a north-to-south "wave" where small plans are moved southward. As this wave moves, it increases the number of small plans whose size is decreasing, so the chance to join small plans increases.

Once the wave is finished, we start another wave running in different direction. We can also use various shapes of waves, e.g., circular. In this case, we choose a car whose plan represents a center of a circle which runs inward. We run various waves as long as our solution is improving.

\subsection{Complexity and scalability}

Theoretically, time and space complexity of our algorithm can be exponential because of the Bin packing heuristic.  However, our experiments show that finding more that 1000 combinations of exchanges for two plans is very rare and such cases can be easily handled by reducing the number of combinations by keeping promising ones only. Furthermore, it is not possible to increase the limit $L$ for larger networks since it decreases the quality of maintenance. Therefore, we can assume that the time and space complexity of the Bin packing heuristic is constant.

Space complexity of the algorithm is linear in the size of the road network since we need to remember a fixed amount of information per road and a list of traversed roads for each car. Clearly, for every car $c_1$ there are only few cars $c_2$ such that plans of $c_1$ and $c_2$ share a common vertex (for planar road network and a constant $L$). Therefore, the Bin packing heuristic is applied linearly many times during every wave of the global algorithm. The number of waves depends on a terminal condition.

Summarising, our algorithm is well-scaleable even for large networks.

\section{Lower bounds} \label{sec:lower}

In this section, we present various ways to prove lower bounds for the minimal number of cars and deadheads needed to maintain all roads as required. The first trivial lower bound on the number of cars is obtained by dividing the total length of all roads by the limit for one car, formally $\left\lceil \frac{\sum_{e \in e} l_e}{L} \right\rceil$. This trivial lower bound can be improved e.g. by including necessary deadheads to ensure that every road is reachable from a depot with the appropriate storage.

\subsection{Connectivity}

Next, we find a set of roads that has to be traversed to ensure that every road is reachable from a depot with the appropriate storage. We find a set $S \subset E_c$ of chemical roads with the minimal total length such that every inert road can be reached from a depot with inert storage by edges of $E_i, E_s$ or $S$. Note that the symmetric case where inert and chemical are exchanged is also relevant.

We describe an integer linear programming problem which calculates these deadheads. We use a binary variable $x_e$ for every chemical road $e \in E_c$ to determine whether $e$ is traversed by an inert car.
\begin{equation}\begin{array}{*{20}{r}} \label{eq:separate}
\text{Minimize} & \sum_{e \in E_c} l_e x_e \\
\text{subject to}
& \sum_{e \in \delta_G R} x_e &\ge& 1 \\
\end{array}\end{equation}
The last inequality is included for every subset of vertices $R \subseteq V$ such that $R$ contains no inert depot and $G[R]$ contains an inert road and $\delta_G R$ contains only chemical roads. Such sets $R$ form cuts between inert roads and inert depots and these cuts have to be traversed by at least one chemical road.

The advantage of this method is a small number of variables (at most $|E|$) but the disadvantage is possibly exponential number of constrains. We solve this issue using Lazy Constraint (see e.g. \cite{STMKG}) as follows. We start with no subset $R$ and solve \eqref{eq:separate}. When an integer linear programming solver finds a solution $x$, we use a graph search algorithm (e.g. breath first search) to find components of the subgraph on edges $E_i \cup E_s \cup \{e \in E_c;\; x_e = 1\}$. For every component on vertices $R$ which contains an inert road but no inert depot, we include the subset $R$ into the problem \eqref{eq:separate}. If there is no such a component, we terminate since every inert road is reachable from an inert depot using roads of $E_i \cup E_s \cup \{e \in E_c;\; x_e = 1\}$. Furthermore, we can use solver's callback to provide these lazy constraints during branch-and-bound process.

There are simple heuristics which significantly reduce the computation time. First, let $S$ be the set of vertices incident with an inert road. Our goal is to find a variant minimal Steiner tree on set $S$ in which the length of inert and snowplow roads is excluded from the objective function and every component of the Steiner tree contains an inert depot. Therefore, we can contract every inert and snowplow road. If we contract two vertices such that at least one of them belongs to $S$ then we add the new vertex into $S$. We also merge all inert depots into a single vertex which is added into $S$. After all contractions, we delete all loops and parallel roads where the shortest road is kept. This simple process reduces the original problem into a classical Steiner tree problem which we solve using a similar iterative method as \eqref{eq:separate}.

We apply the following heuristics which reduces a realistic road network further.
\begin{enumerate}
\item Remove vertices $u$ of degree 1. If $u \in S$, then the only neighbour $v$ of $u$ has to be added into $S$ and the length $uv$ has to be added to the optimal solution.
\item Remove vertices $u \notin S$ of degree 2 as follows. Let $v$ and $w$ be neighbors of $u$. Replace roads $uv$ and $uw$ by a new road $uw$ of length $l_{uv} + l_{uw}$. Again, only the shortest edges of parallel ones is kept.
\item If the graph contains a cycle with a road $e$ of length larger than the sum of lengths of all other roads on the cycle, then we can remove $e$.
\end{enumerate}

\subsection{Snowplow}

These methods find the minimal deadheads for inert and chemical cars separately without any attention on snowplow roads. For example, it may happen that a snowplow road has to be traversed from a depot to reach a crossroad with chemical and inert roads. In this case, there is a deadhead on the snowplow road either by a chemical or an inert car, but this deadhead is not counted in previous lower bounds.

For every road $e \in E$ we consider a binary variable $x_e$ which equals to one if there is a deadhead on $e$, and for every snowplow road $e \in E_s$ we consider a binary variable $x^m_e$ which equals to one if $e$ is maintained by a chemical car. Note that for a snowplow road $e$, if $x_e = 1$ then $e$ is traversed by both chemical and inert cars, so $x^m_e$ can have either value. We implemented the following linear programming problem.

\begin{equation}\begin{array}{*{20}{r}} \label{eq:together}
\text{Minimize} & \sum_{e \in E} l_e x_e \\
\text{subject to}
& \sum_{e \in \delta_G R^c} x_e & + x^m_e &\ge& 1 \\
& \sum_{e \in \delta_G R^i} x_e & + 1 - x^m_e &\ge& 1 \\
\end{array}\end{equation}

Similarly as in \eqref{eq:separate}, we consider every subset of vertices $R^c \subseteq V$ such that $R^c$ contains no chemical depot and $G[R^c]$ contains a chemical road and $\delta_G R^c$ contains no chemical road; and analogously for subset $R^i$ in the inert case. Note that we cannot easily reduce \eqref{eq:together} into a Steiner tree problem since it is not possible to contract all chemical, inert nor snowplow roads. Furthermore, heuristics for model \eqref{eq:separate} cannot be applied for \eqref{eq:together} in general.

\subsection{Multiple cars}

Method \eqref{eq:together} is equivalent to minimizing deadheads under the condition that every depot has a chemical and an inert car without capacity limit $L$ if the depot has the appropriate storage. Now, we incorporate the limit $L$.  For simplicity assume that sets $C_c$ of chemical cars and $C_i$ of inert cars are given. Our goal is to determine whether $|C_c|$ chemical cars and $|C_i|$ inert cars is sufficient and minimize deadhead if possible.

For every car $c \in C$ and every road $e \in E$ we consider a binary variable $x^c_e$ which determines whether $e$ is traversed by $c$.
\begin{equation}\begin{array}{lrcll} \label{eq:multiple-cars}
\text{Minimize} & \sum_{c \in C} \sum_{e \in E} l_e x_e \\
\text{subject to}
& \sum_{c \in C_c} x^c_e &\ge& 1 & \text{for every $e \in E_c$} \\
& \sum_{c \in C_i} x^c_e &\ge& 1 & \text{for every $e \in E_i$} \\
& \sum_{c \in C} x^c_e & \ge& 1 & \text{for every $e \in E_s$} \\
& |E(G[R])| \sum_{e \in \delta_G R} x^c_e &\ge& \sum_{e \in E(G[R])} x^c_e & \text{for every $c \in C$ and $R$} 
\end{array}\end{equation}
In this integer linear programming problem, we ensure that every road is traversed by at least one car of the appropriate maintaining method. The last inequality is applied for every car $c \in C$ and every subset of vertices $R \subseteq V$ which does not contain a depot with the appropriate storage. This inequality ensures that if $c$ traverses a road of the subgraph $G[R]$, then $c$ traverses at least one road of the cut $\delta_G R$.

However, this approach is not suitable for realistic scales. There are exponentially many constrains which we also tried to handle by iterative generation. But there too many variables (in the order of $|E||C|$) and the coefficient $|E(G[R])|$ significantly slows down solving the integer problem. Another problem is that for every solution there are $|C_i|! \cdot |C_c|!$ symmetric solutions obtained by relabeling cars. In order to avoid these symmetries, we can order all chemical roads and cars and add linear constrains for every $i$ and $j$ which ensure that $i$-th chemical car traverses at least one of $1, \ldots, j$ chemical roads or $j$-th chemical road is traversed by at least one of $1, \ldots, i$ chemical car; and similarly for inert. But this modification enlarges \eqref{eq:multiple-cars} by about $|C||E|$ constrains with $|C||E|^2$ non-zero coefficients.

\section{Case study}

\begin{figure}[ht]
 
\begin{subfigure}{0.5\textwidth}
\includegraphics[width=0.9\linewidth]{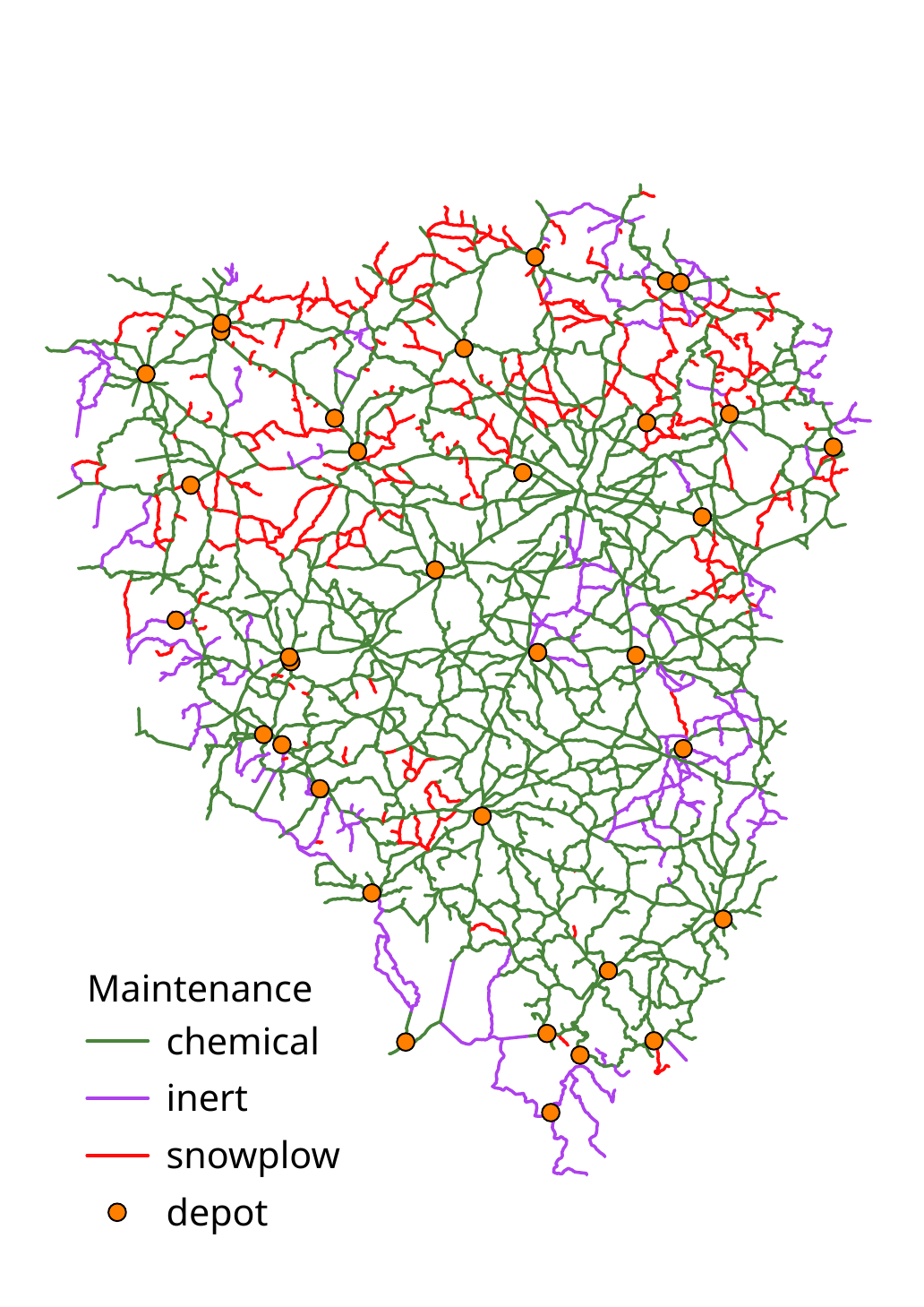} 
\caption{Method of maintenance}
\label{fig:material}
\end{subfigure}
\begin{subfigure}{0.5\textwidth}
\includegraphics[width=0.9\linewidth]{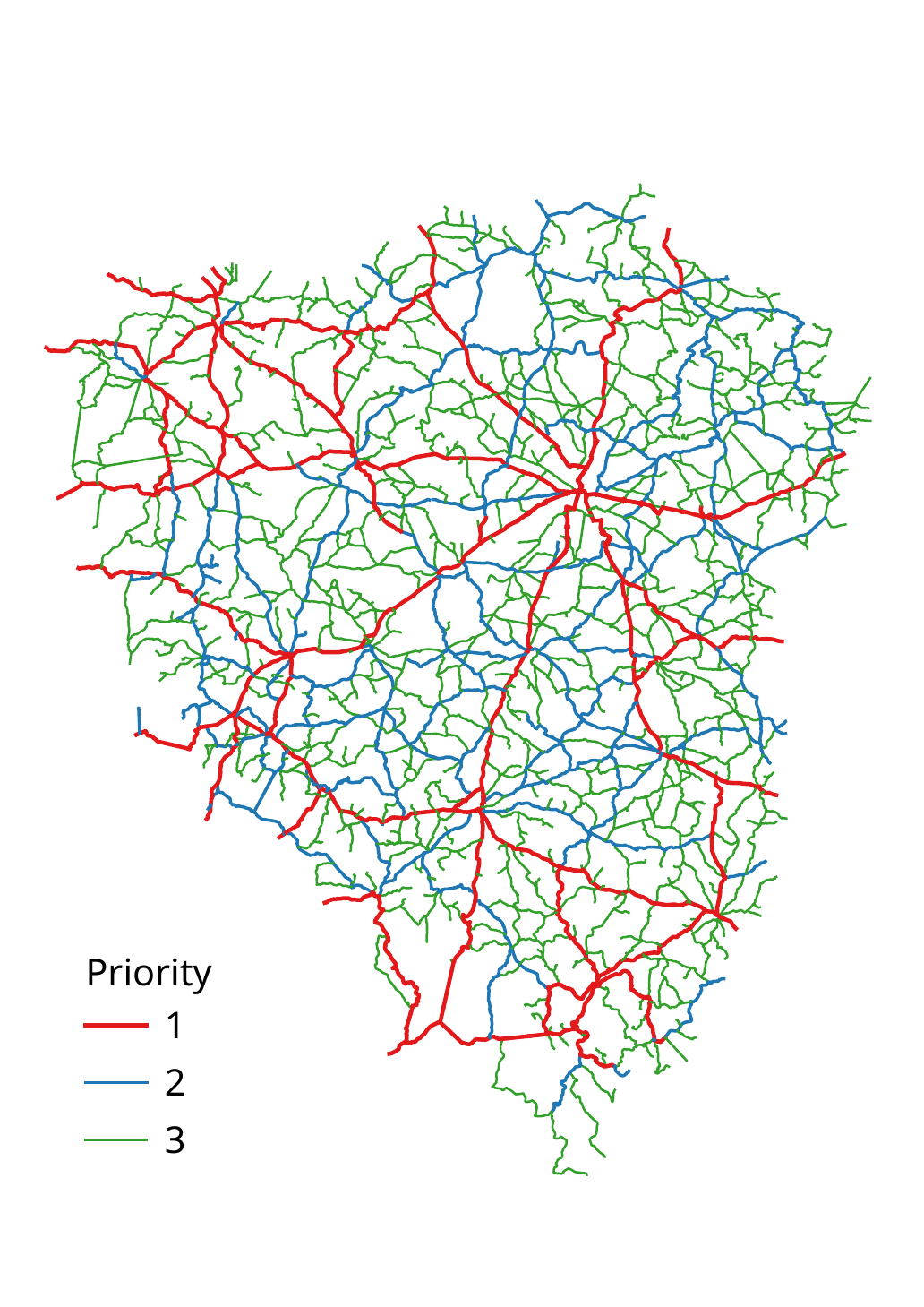}
\caption{Priority levels}
\label{fig:priority}
\end{subfigure}
 
\caption{Maintenance methods and priority levels of the road network}
\label{fig:material_and_priority}
\end{figure}

In order to verify our algorithm, we created a realistic road network on 1719 vertices and 2280 edges of total length 4860 km. Table \ref{tab:type_of_maintenance} shows length of roads for each priority and maintenance methods. Figure \ref{fig:material_and_priority} gives geographical overview of the road network. We can see in figure \ref{fig:material} that roads maintained by inert are concentrated in several areas representing nature parks and water sources while remaining roads are maintained chemically except some snowplowed dead-end road. Furthermore, the priority in figure \ref{fig:priority} shows how high-priority roads form main transit system while low-priority roads complete the network to attach small villages.

Crucial 34 vertices of the road network were selected for depots, 4 of them have only chemical storage, 6 have only inert storage and 24 have both storages.

\begin{table}[h]
    \centering
    \begin{tabular}{|c|r|r|r|r|r|}
    \hline
    {\bf maintenance/priority} & 1 & 2 & 3 & {\bf sum}\\
    \hline
    {chemical} & 840 & 1080 & 1472 & 3392 \\
    {inert} & 42 & 65 & 569 & 676 \\
    {snowplow} & 8 & 15 & 724 & 747 \\
    {\bf sum} & {\bf 890} & {\bf 1160} & {\bf 2765} & {\bf 4815} \\
    \hline
    \end{tabular}
    \caption{Length of roads of each priority and maintenance method [in kilometers]}
    \label{tab:type_of_maintenance}
\end{table}

All roads have to be maintained during 8 working hours in which a car has to be loaded and a driver needs safety breaks. Therefore, we decided that it should be safe to maintain all roads in 6 hours with neither safety breaks nor other auxiliary times, e.g., for loading. Since average speed of a maintaining car is about 30 km/h and every road has to be maintained in both directions, we set the limit $L$ to be 90 km. In order to increase frequency of maintenance of higher priority roads, we decided to maintain the first priority roads three times, the second priority twice and third priority only once. Table \ref{tab:frequency} summarizes the length of roads including frequency of maintenance.

\begin{table}[]
    \centering
    \begin{tabular}{|c|r|r|r|r|r|}
    \hline
    {\bf priority} & 1 & 2 & 3 & {\bf sum}\\
    \hline
    length & 890 & 1160 & 2765 & 4815 \\
    multiplicity & 3 & 2 & 1 & \\
    multiplied length & 2670 & 2320 & 2765 & 7755 \\
    \hline
    \end{tabular}
    \caption{Length of roads including frequency of maintenance [in kilometers]}
    \label{tab:frequency}
\end{table}

\section{Results}

\subsection{Solution found by our algorithm}

We implemented our algorithm in Python 3.7.3 and run the program on a single thread on a laptop with CPU Intel Core i5-7200U at 2.50GHz. The running time needed to find the following solution was 2 minutes and 36 seconds and the program uses 82 MB of memory. We use Python for its rapid prototyping, but a proper implementation in C/C++ should be able to solve the instance in few seconds.

Our program found a solution with 93 cars and 361 km deadheads; see Table \ref{tab:result}. However as we can see in the table, most of deadheads are caused by the inert maintenance method. This is not surprising since as we can see in Figure \ref{fig:material}, roads requiring inert maintenance are distributed so that significant deadheads are necessary to reach them from depots.

\begin{table}[ht]
    \centering
    \begin{tabular}{|c|rr|rr|rr|}
    \hline
    {\bf method} & \multicolumn{2}{c|}{chemical} & \multicolumn{2}{c|}{inert} & \multicolumn{2}{c|}{{\bf sum}}\\
    & km& \% & km & \% & km & \% \\
    \hline
    number of cars & \multicolumn{2}{c|}{78} & \multicolumn{2}{c|}{15} & \multicolumn{2}{c|}{93} \\
    total limit (90 km per car) & 7020 & & 1350 & & 8370 & \\
    maintaining (\% of limit) & 6709 & (95.57) & 1046  & (77.48) & 7755  & (92.65) \\
    deadhead (\% of limit) & 155  & (2.21) & 206  & (15.26) & 361  & (4.31) \\
    unused (\% of limit) & 156  &(2.22) & 98  & (7.26) & 254  & (3.03) \\
    \hline
    \end{tabular}
    \caption{The summary of our solution: there are 93 cars and the length of the plan of each car is at most 90 km. Hence all cars can traverse at most 8370 km of which, in our solution, 7750 km are maintaining, 361 km is deadhead and the remaining 254 km of the upper bound are unused.}
    \label{tab:result}
\end{table}

\subsection{Lower bounds}

For lower bounds, we used a more powerful computer with 8 processors Intel Core i7-6700 CPU at 3.40GHz and 32 GB RAM. Integer linear programming problems were solved using Gurobi 8.0.1.

LP models \eqref{eq:separate} and \eqref{eq:together} are solved efficiently in less than 100 seconds since these models have only $\Theta(|E|)$ variables, see Table \ref{tab:lower}. Heuristics for \eqref{eq:separate} significantly reduce the size of a road network, so the LP model can be solved in 16 seconds. However, \eqref{eq:multiple-cars} requires $\Theta(|C||E|)$ variables and therefore the solver run out of memory. 

Using LP model \eqref{eq:separate} we calculate that at least 143 km of chemical roads have to be traversed by inert cars to reach all inert roads from depots. Similarly, at least 7 km of inert roads has to be traversed by chemical cars, so the total deadheads need to be at least 150 km. This analysis confirms our expectation and results presented in Table \ref{tab:result} that inert cars have more deadheads eventhough the total length of inert roads is significantly smaller than of chemical roads. LP model \eqref{eq:together} increases the lower bound on deadheads by 467 m only which is expected since cases where a snowplow road has to be traversed by both a chemical and inert cars are very rare.

Our solution has 361 km of deadheads of which 150 km is caused by disconnectivity. Reasons for remaining 211 km of deadheads are:
\begin{itemize}
\item leaving depots (degree of a vertex with a depot may be smaller than the number of cars),
\item priority rule (roads of lower priority have to be traversed to reach a road of higher priority), and
\item suboptimality of our solution.
\end{itemize}

Note on Table \ref{tab:result} that 254 km are unused which is mainly caused by the fact that a single road cannot be split and maintained by two or more cars, so it is not possible to find a subset of roads of total length exactly 90 km. Furthermore, chemical cars do not use 2.22 \% of the limit and inert cars 7.26 \%. Here, inert is less efficient since sparsity of inert roads cause smaller number of combinations.

Since the total maintaining length of roads is 7755 km and each car can maintain at most 90 km, at least 87 cars are necessary. When the maintaining length is increased by 150 km of provably unavoidable deadheads, the optimal number of cars is at least 88.

\begin{table}[ht]
    \centering
    \begin{tabular}{|l|rrr|}
    \hline
    Methods & Vertices & Roads & Time \\
    \hline
    \multicolumn{4}{|c|}{\textbf{Chemical cars using inert roads \eqref{eq:separate}}} \\
    \hline
    No heuristic & 1716 & 2271 & $< 1$ s \\
    Reduction to Steiner tree & 154 & 191 & $< 1$ s \\
    Further heuristics & 14 & 22 & $< 1$ s \\
    \hline
    \multicolumn{4}{|c|}{\textbf{Inert cars using chemical roads \eqref{eq:separate}}} \\
    \hline
    No heuristic & 1716 & 2271 & 96 s \\
    Reduction to Steiner tree & 1124 & 1599 & 75 s \\
    Further heuristics & 639 & 1028 & 16 s \\
    \hline
    \multicolumn{4}{|c|}{\textbf{Join chemical and inert cars \eqref{eq:together}}} \\
    \hline
    No heuristic & 1716 & 2271 & 99 s \\
    \hline
    \end{tabular}
    \caption{Number of vertices and roads with computational time of lower bounds techniques.}
    \label{tab:lower}
\end{table}

\section{Conclusion}

This paper introduces a new variant of arc routing problem with addition of practical constrains including different maintenance methods and priorities of roads and presents an algorithm for large-scale instances. This algorithm is able to give very good results in few minutes even on an ordinary laptop.

However, we believe it is possible to obtain better solutions using more sophisticated algorithms. For instance, our algorithm is based on local search, so it may find just a local optima. It may be possible to better explore the solution space if our algorithm is combined with Artificial Intelligence, e.g. genetic algorithms.

An interesting and practically significant modification of our problem is a real-time variant where weather conditions are changing continuously and the task is to maintain the road network as well as possible. In this case, our fast heuristic could be used for on-line planning.

\bibliographystyle{amsplain} %AMS bibliography style

\end{document}